\documentclass[10pt]{article} 
\usepackage[accepted]{tmlr}

\usepackage{hyperref}
\usepackage{url}

\usepackage{amsmath}
\usepackage{amsfonts}
\usepackage{acro}
\usepackage{pgfplots}
\usepgfplotslibrary{groupplots,dateplot}
\usepackage{soul}
\usepackage{tikz}
\usepackage{bm}
\usepackage{booktabs}
\usepackage{threeparttable}
\usepackage{algorithm}
\usepackage[noend]{algpseudocode}
\usetikzlibrary{arrows,arrows.meta,patterns,positioning,shapes}
\usepackage{enumitem}
\usepackage{siunitx}
\usepackage{multirow}
\usepackage{adjustbox}
\usepackage{csvsimple}
\usepackage[capitalise]{cleveref}

\usepackage{xspace}

\DeclareSIUnit{\pp}{\textup{p.p.}}

\newcommand{\cocoflf}{$\text{CoCoFL}^{\text{F}}$}
\newcommand{\cocoflff}{$\text{CoCoFL}^{\text{FF}}$}
\newcommand{\cocoflqff}{$\text{CoCoFL}^{\text{QFF}}$}

\DeclareAcronym{NLP}{short=NLP, long=natural language processing,short-indefinite=an}
\DeclareAcronym{CNN}{short=CNN,long=convolutional neural network}
\DeclareAcronym{CL}{short=CL, long=convolutional layer}
\DeclareAcronym{FC}{short=FC,long=fully-connected}
\DeclareAcronym{FLOPS}{short=FLOPS,long=floating point operations per second}
\DeclareAcronym{IOT}{short=IoT,long=internet of things,short-indefinite=an}
\DeclareAcronym{LUT}{short=LUT,long=lookup table}
\DeclareAcronym{MAC}{short=MAC,long=multiply-accumulate operation}
\DeclareAcronym{FL}{short=FL,long=federated learning,short-indefinite=an}
\DeclareAcronym{NN}{short=NN,long=neural network,short-indefinite=an}
\DeclareAcronym{FEDAVG}{short=FedAvg, long=federated averaging}
\DeclareAcronym{IID}{short=iid, long=independent and identically distributed, short-indefinite=an}
\DeclareAcronym{ML}{short=ML, long=machine learning, short-indefinite=an}
\DeclareAcronym{SGD}{short=SGD, long=stochastic gradient descent}
\DeclareAcronym{BN}{short=BN, long=batch normalization}
\newcommand{\bftab}{\fontseries{b}\selectfont}
\pgfplotsset{
    compat=1.15,
}
\usetikzlibrary{plotmarks}
\usepgfplotslibrary{fillbetween}
\usetikzlibrary{arrows, decorations.markings, decorations}

\definecolor{tabblue}{rgb}{0.121, 0.466, 0.705}
\definecolor{taborange}{rgb}{1, 0.498, 0.054}
\definecolor{tabgreen}{rgb}{0.172, 0.627, 0.172}
\definecolor{tabred}{rgb}{0.827, 0.192, 0.203}

\usepackage{subcaption}

\title{CoCoFL: Communication- and Computation-Aware Federated Learning via Partial NN Freezing and Quantization}

\author{\name Kilian Pfeiffer \email kilian.pfeiffer@kit.edu \\
      \addr Karlsruhe Institute of Technology
      \AND
      \name Martin Rapp \email martin.rapp@kit.edu \\
      \addr Karlsruhe Institute of Technology
      \AND
      \name Ramin Khalili \email ramin.khalili@huawei.com\\
      \addr Huawei Research Center Munich
      \AND
      \name Jörg Henkel \email henkel@kit.edu\\
      \addr Karlsruhe Institute of Technology}

\def\month{06}  
\def\year{2023} 
\def\openreview{\url{https://openreview.net/forum?id=XJIg4kQbkv}} 

\begin{document}

\maketitle

\begin{abstract}
    Devices participating in \ac{FL} typically have heterogeneous communication, computation, and memory resources. However, in synchronous \ac{FL}, all devices need to finish training by the same deadline dictated by the server. Our results show that training a smaller subset of the \ac{NN} at constrained devices, i.e., dropping neurons/filters as proposed by state of the art, is inefficient, preventing these devices to make an effective contribution to the model. This causes unfairness w.r.t the achievable accuracies of constrained devices, especially in cases with a skewed distribution of class labels across devices.
We present a novel FL technique, \emph{CoCoFL}, which maintains the full NN structure on all devices. To adapt to the devices' heterogeneous resources, CoCoFL freezes and quantizes selected layers, reducing communication, computation, and memory requirements, whereas other layers are still trained in full precision, enabling to reach a high accuracy. Thereby, CoCoFL efficiently utilizes the available resources on devices and allows constrained devices to make a significant contribution to the \ac{FL} system, preserving fairness among participants (accuracy parity) and significantly improving final accuracy.
\end{abstract}

\acresetall
\section{Introduction}
\label{sec:intro}

Deep learning has achieved impressive results in many domains~\citep{he2015deep,densenet, young2018recent}, and is also being applied in embedded systems such as mobile phones or \ac{IOT} devices~\citep{dhar2019device}.
With recent hardware improvements, these devices are not only capable of performing inference of a pre-trained model, but also of on-device training. Hence, \ac{FL}~\citep{mcmahan2017communicationefficient} has emerged as an alternative to central training. 
\Iac{FL} system comprises many \emph{devices} that each train a deep \ac{NN} on their private data, and share knowledge by exchanging \ac{NN} parameters via a \emph{server}.
Distributing learning through \ac{FL} brings many benefits, most importantly preserving the privacy of the end users.

Devices in real-world systems have limited computation, communication, and memory resources for training, varying across devices. For instance, smartphones that participate in \iac{FL} system have different performance and memory (e.g., different hardware generations), and the conditions of their wireless communication channels vary (e.g., due to fading~\citep{goldsmith_2005}).
Similar observations can be made in \ac{IOT} systems~\citep{bhardwaj2020new}.
As stated by prior art~\citep{rapp2022distreal,xu2021helios,diao2020heterofl,horvath2021fjord}, to enable efficient learning in such systems, \ac{FL} needs to adapt to the per-device constraints, i.e., \emph{hardware-aware \ac{FL}}. Although different techniques are proposed, the common idea in these state-of-the-art solutions is to reduce the complexity by training subsets of the \ac{NN} model on less capable devices, to match the required resources for training to the actual resource availability.
While these techniques enable constrained devices to participate in the training, they do not effectively learn from their data, i.e., they do not preserve fairness (accuracy parity~\citep{fairness}).
This is especially critical with non \acl{IID} (non-\acs{IID}) data, where the data differs statistically between devices~\citep{hsu2019measuring}.
Our evaluation results show that existing solutions perform poorly in non-\ac{IID} cases, such that in some settings, simply excluding constrained devices from training reaches higher accuracies.
We attribute this in part to the fact that updates at constrained devices are less relevant to the overall learning objective, as they train much smaller subsets of the model, and also to the inability of these solutions to efficiently use the available resources in constrained devices.
\begin{figure*}
    \centering

    \trimbox{0.15cm 1.5cm 0.20cm 0.2cm}{
    \includegraphics[page=1]{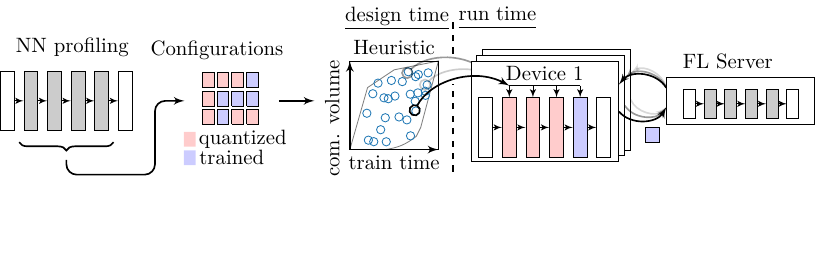}
    }
    
    \caption{
        Overview of CoCoFL.
        At design time, different configurations of frozen/trained layers are profiled w.r.t.\ communication, computation, and memory in training.
        At run time, a heuristic selects a suitable configuration on each device w.r.t.\ the device's constraints.
    }
    \label{fig:system_model}
\end{figure*}

In this paper, we propose a new technique CoCoFL (\cref{fig:system_model}), that allows all devices to calculate gradients based on the full model, irrespective of their capabilities, through partial freezing and quantization of the model at constrained devices.
We show that \emph{quantizing frozen layers but keeping trained layers at full precision} results in a large reduction in resource requirements, while still enabling efficient learning at devices. This combination has not been exploited so far.
Freezing layers reduces the required gradient computations, the storage of intermediate activations, and the size of the parameter update, while quantization further speeds up the computations of frozen layers.
Thereby, our solution adjusts the complexity of training to the resources available at each device.
Partial freezing and quantization opens up a large design space, where each layer can be frozen or trained on each participating device.
The selection of trained layers has a significant impact on the required resources and on the accuracy.
We introduce a heuristic that allows for server-independent selection of layers w.r.t.\ local resource availability at run time, based on design-time profiling of the performance of devices.
\emph{We demonstrate that our solution reaches significantly higher accuracy in \ac{IID} and non-\ac{IID} data, when compared with the state of the art, significantly improving \ac{FL} systems.}

In summary, we make the following novel contributions:
\begin{itemize}[noitemsep, nolistsep]
\item We empirically show that in many scenarios, state-of-the-art subset-based techniques do not reach better accuracies than simply excluding less capable devices (a straightforward baseline). We observe this throughout various datasets (e.g., CIFAR10, XChest, and Leaf benchmark data), data distributions, and \ac{NN} topologies (e.g., ResNet, DenseNet, and Transformers).
\item Compared to the state of the art, in these scenarios, we enable increased fairness of contribution and higher final accuracies in \ac{FL} with heterogeneous resources by allowing less capable devices to do training based on the full \ac{NN} structure. This is achieved by the following technical contributions.
\item We introduce a novel partial freezing and quantization technique to adjust to computation, communication, and memory constraints of devices that allows to train full layers of \acp{NN}.
\item We introduce \emph{CoCoFL}\footnote{The code is available at \url{https://github.com/k1l1/CoCoFL}.}, based on partial freezing and quantization, with a simple, yet effective, heuristic to select locally on each device which layers to freeze or train based on the available communication, computation, and memory resources.
\end{itemize}

\section{System Model and Problem Definition}
\label{sec:model}

\textbf{System Model:}
We target a distributed system comprising a \emph{server} that is responsible for coordination
and \emph{devices}~$\mathcal  C$ that act as clients.
Each device~$c\in\mathcal C$ has exclusive access to its local data~$\mathcal D_c$.
Training is done iteratively using \ac{FL} in synchronous rounds~$r$.
In each round, a subset of the devices~$\mathcal C^{(r)}\subset \mathcal C$ is selected. Each selected device downloads the latest model parameters~$w^{(r)}$ from the server, performs training on its local data for a pre-defined \emph{round time}~$T$, and then uploads the updated model parameters to the server.
The server averages all received updates (FedAvg~\citep{mcmahan2017communicationefficient}) to build~$w^{(r+1)}$ for the next round:
\begin{equation}
    \textstyle
    w^{(r+1)} = \frac{1}{\sum_{c \in \mathcal C^{(r)}} |\mathcal{D}_c|} \cdot
    \sum_{c \in \mathcal C^{(r)}} |\mathcal D_c| \cdot w_c^{(r)} \label{eq:fedavg}
\end{equation}
The server discards updates that arrive late (straggler), i.e., devices must upload their updates in time.

\textbf{Device Model:}
Devices are \emph{heterogeneous w.r.t.\ their computation (performance), memory, and communication constraints}.
The performance of a device (how long the training of \iac{NN} takes) depends on its hardware (number of cores, microarchitecture, memory bandwidth, etc.), software (employed deep learning libraries, etc.), and training configuration (topology, amount of data, etc.).
Similarly, the available memory~$M_c$ of device~$c$ depends on its hardware, while the required memory during training depends on the software and training configuration.
Some of these configurations are fixed by the \ac{FL} system (\ac{NN} topology, etc.), while others are fixed by the device (hardware, software, amount of data), but some configuration~$A_c^{(r)}$ can be adjusted per device per round.
In our case, $A_c^{(r)}\in\mathcal A$ describes the subset of all \ac{NN} layers that are trained in round~$r$ by device~$c$, with~$\mathcal A$ being the set of all configurations (see \cref{sec:technique_block_selection}).
The training time of device~$c$ for any $A\in\mathcal A$ is represented by the function~$t_c : \mathcal A{\to}\mathbb{R}$.
The required memory during training is represented by $m_c : \mathcal A{\to}\mathbb{R}$.
We obtain~$t_c$ and $m_c$ through profiling our technique on real hardware (measuring the training time and peak memory usage for different~$A$).

The communication channel between devices and servers is commonly asymmetric: The download link from the server to the devices can be neglected due to the commonly high transmit power of base stations~\citep{yang2020energy}. The upload link from devices to the server is subject to heterogeneous channel quality, as discussed in \cref{sec:intro}.
Therefore, we model the communication constraint~$S_c^{(r)}$ of a device~$c$ in round~$r$ as a limit in the number of bits that can be uploaded to the server at the end of the round.
In our case, all layers not contained in $A_c^{(r)}$ are frozen (and quantized).
Their parameters do not change, hence, do not need to be uploaded to the server.
We represent the size of the parameter update for any $A\in\mathcal A$ by a function $s : \mathcal A{\to}\mathbb{N}$ that is independent of device characteristics.
Function~$s$ can be derived analytically or by counting parameters per layer.

\textbf{Problem Definition:}
Our main objective is to maximize the \emph{final accuracy}~$acc$ of the server model~$w^{(R)}$ after $R$~rounds under communication, computation, and memory constraints, by selecting per-device per-round the set of trained layers~$A_c^{(r)}$:
\begin{equation}
    \text{maximize}~acc(w^{(R)}) \quad  \forall c\in\mathcal C \quad \forall 1\le r\le R\\ \quad \quad 
    \text{s.t.}\ t_c(A_c^{(r)}){\le}T\
    \land\ m_c (A_c^{(r)}){\leq}M_c\ \land\ s(A_c^{(r)}){\le}S_c^{(r)}
    \label{eq:constraint}
\end{equation}
We also evaluate fairness (accuracy parity~\citep{fairness}) as a secondary metric, by measuring the \emph{device- or group specific accuracy} using data that reflects each device's or group's distribution of local data $\mathcal D_c$. 

\section{Related Work}

We divide the related work into works that employ a similar mechanism (quantization/freezing) and works that target a similar problem (computation/communication/memory constraints in \ac{FL}).

\textbf{Quantization and Freezing in Centralized Training:}
Most works on quantization target the inference, with full-precision training.
Naive \emph{training} on quantized parameters leads to training stagnation as small gradients are rounded to zero~\citep{li2017training}.
To solve this, one branch of works performs stochastic rounding~\citep{stochastic_rounding}.
However, stochastic rounding prevents convergence to the local minimum in the final phase of training with a low learning rate~\citep{li2017training}, reducing the accuracy.
Another branch of work uses a full-precision copy of the parameters as an accumulator~\citep{micikevicius2018mixed}.
Calculating the parameter gradient based on quantized activations and parameters induces instability to the learning processes, which requires a lower learning rate to maintain convergence, slowing down the training, but also still resulting in a lower accuracy~\citep{guo2018survey}.
\emph{In summary, achieving fast convergence and high accuracy requires keeping the trained layers in full precision (activation and parameters in the forward and backward pass).}
\cite{goutam2020layerout} stochastically freeze layers of \iac{NN} to speed up training, keeping the frozen layers at full precision which limits the achievable speedup.
Also, due to its stochastic nature, it is not applicable to a hard computation constraint.
All these works either apply quantization or freezing.
\emph{None of the existing works has exploited the symbiosis between freezing and quantization, where only frozen layers are quantized to maintain good convergence properties.}

\textbf{Communication, Computation, and Memory Constraints in FL:}
Most works on resource-constrained \ac{FL} have targeted communication, extensively studying compression, quantization, and sketching of parameter updates ~\citep{shi2020communication,thakker2019compressing}.
All perform regular training of the full \ac{NN} in full precision, requiring full computation and memory resources, and only reduce the size of the parameter update.
They are orthogonal to ours, i.e., applicable on top of CoCoFL.
The work in~\cite{chen2021communication} detects \ac{NN} parameters that have stabilized, freezes them, and excludes them from synchronization to reduce the required communication.
However, this technique can not cope with a given communication constraint.
Recently, a preliminary work~\citep{yang2021partial} proposed freezing layers to save communication and memory in \ac{FL}.
It exploits that frozen layers do not require storing activations for computing gradients, and do not need to be uploaded to the server.
This technique has later been combined with quantization during download and upload~\citep{ro2022scaling}, but unlike in our CoCoFL, quantization is not used during training, missing out on significant optimization opportunities (e.g., up to $4\times$ lower computation time as we will show in our experiments).
All these works do not reduce the computation cost of training.

Computation constraints in \ac{FL} devices have only recently attracted attention.
Some employ asynchronous \ac{FL}~\citep{xie2020asynchronous}.
However, this does not reduce memory requirements and may reduce the convergence stability~\citep{mcmahan2017communicationefficient}.
FedProx~\citep{li2020federated} dynamically drops training data on straggling devices, which reduces computations but does not affect communication and memory requirements, and reduces the contribution of less capable devices.
DISTREAL~\citep{rapp2022distreal} employs dropout to dynamically reduce the size of the trained \ac{NN}, reducing computations.
They still transmit the full \ac{NN} updates, and, hence, do not reduce communication costs.
Several others train subsets of the \ac{NN} on each device by (temporarily) scaling the width (number of filters/neurons) of layers.
This may save communication, computation, and memory.
In particular, Helios~\citep{xu2021helios}, HeteroFL~\citep{diao2020heterofl}, and FjORD~\citep{horvath2021fjord} proposed to create a separate subset per each device according to its available resources.
However, training very small subsets on weak devices does not enable them to effectively learn from their data, as we show in our evaluation, reducing fairness, hence, the final accuracy.
Additionally, the reductions in communication and computation achieved by width scaling are tightly coupled. Consequently, one of them forms the bottleneck, resulting in unexploited resources in the other metric.
Finally,~\cite{yang2020energy} study the trade-off between communication and computation to minimize the overall energy consumption.
This work is not applicable to a per-device computation or communication constraint.

\emph{In summary, none of the existing works on resource-con\-strained \ac{FL} can adapt to per-device communication, computation, and memory constraints, while still effectively learning from all data on all devices.
We achieve this through our novel combination of freezing and quantization.}

\begin{figure*}
    \centering
    \trimbox{0.0cm 0.1cm 0.2cm 0.0cm}{ 
    \centering
    \includegraphics[page=1]{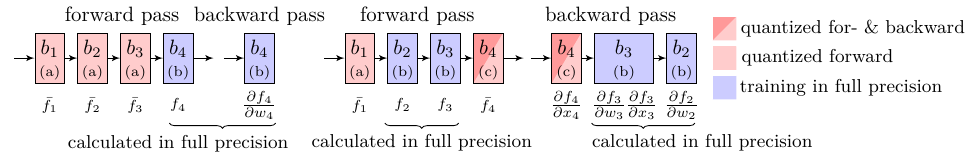}
    }
    \caption{
        The left part shows \iac{NN} where only block $b_4$ is trained, while all others are frozen, and the forward passes of $b_1, b_2$, and $b_3$ are quantized ($\bar f_1, \bar f_2$, and $\bar f_3$), not requiring a backward pass.
        The right part shows two blocks $b_2,b_3$ being trained, therefore requiring intermediate gradients of the quantized block $b_4$. Blocks are labeled (a), (b), and (c) depending on their types.
    }
    \label{fig:blocks_technique}
\end{figure*}

\section{Partial Freezing and Quantization}
\label{sec:technique}
This section introduces our freezing, layer fusion, and quantization technique to reduce the training complexity, which will be used in the CoCoFL algorithm (\cref{sec:technique_block_selection}).
\subsection{Background: \ac{NN} Structure and Training}
\textbf{Structure:}
Common state-of-the-art deep \acp{NN} like ResNet~\citep{he2015deep}, DenseNet~\citep{densenet}, or MobileNet~\citep{howard2017mobilenets} follow a similar structure, where \iac{CL} is followed by \iac{BN} layer, followed by ReLU activation.
These layers account for the majority of training time.
We label repeating structures like this a \emph{block}, where in a general case, \iac{NN} comprises~$N$ blocks. We treat blocks as the smallest entity that is either frozen or trained.
Note that our technique is also applicable to variants of \acp{NN}, (e.g., multiple skip connections or Transformers as we show in our evaluation) but for the sake of simplicity, the block description follows the ResNet structure.

\textbf{Training:} An update step in \acp{NN} comprises a forward and a backward pass. We describe the forward pass as a chain of consecutive operations, where each block~$i$ has an associated forward function~$x_{i+1} = f_i(x_i)$ with parameters~$w_i$. The full forward pass of the \ac{NN} is calculated as~$\hat y = f_{N}(f_{N-1}(\ldots(f_1(x_1)))$, where~$\hat y$ is the \ac{NN}'s output and~$x_1$ is the input.
The backward pass consists of several gradient calculations to compute the parameter gradients~$\frac{\partial E}{\partial w_i}$ for each block~$i$, where~$E =  \mathcal L(\hat y, y)$ is the optimization criterion with some loss function~$\mathcal L$. Using the chain rule, the calculations can be split into several gradient calculations.
In the general case, the gradients w.r.t.\ a block's parameters can be expressed as 
\begin{equation}
\frac{\partial E}{\partial w_i} = \frac{\partial E}{\partial \hat y} \Big(\prod\nolimits_{k=i+1}^{N} \frac{\partial f_{k}(x_{k})}{\partial x_{k}}\Big)
\frac{\partial f_{i}(x_{i})}{\partial w_{i}} \label{eq:backprop_chain_rule}.
\end{equation}
Using the calculated gradients $\frac{\partial E}{\partial w_i}$, local training with \ac{SGD} obtains updated parameters~$\tilde w_i = w_i {-} \eta \frac{\partial E}{\partial w_i}$, where~$\eta$ is the learning rate.
When calculating gradients of several blocks, intermediate gradient computations can be reused.
\subsection{Freezing, Fusion, and Quantization of Blocks}
\textbf{Freezing:}
Freezing a parameter removes the need to calculate its gradients. As by \cref{eq:backprop_chain_rule}, the number of required intermediate gradients depends on the block's index (e.g., the calculation of~$\frac{\partial E}{\partial w_N}$ requires no intermediate gradients, while~$\frac{\partial E}{\partial w_1}$ requires intermediate gradients from all other blocks).
Based on the required per-block operations, we distinguish between three block types (illustrated in \cref{fig:blocks_technique}):
{
\setlength\leftmargini{2em}
\begin{enumerate}[label=(\alph*), noitemsep, nolistsep]
    \item \emph{Frozen block:} With no preceding trained block, a frozen block only requires a forward pass~$f_i(x)$.
    \item \emph{Trained block:} Trained blocks require a forward pass~$f_i(x_i)$, calculation of gradients w.r.t.\ their parameters~$\frac{f_i(x_i)}{\partial w_i}$, and gradients w.r.t the input~$\frac{\partial f_i(x_i)}{\partial x_i}$ for preceding trained blocks.
    \item \emph{Frozen block with backward pass:} With preceding trained blocks, a frozen block requires the forward pass~$f_i(x_i)$ and intermediate gradients w.r.t.~the input~$\frac{\partial f_i(x_i)}{\partial x_i}$.
\end{enumerate}
}
\emph{Consequently, freezing blocks reduces the number of per-block operations of frozen blocks (from~3 to~2 or~1), and therefore saves \acp{MAC}, reducing computation time.} Similarly, if a layer is not trained, the activation values~$x_i$ can be released in memory during the forward pass, reducing the memory footprint. Additionally, parameters of frozen layers do not change throughout an \ac{FL} round, therefore, do not have to be uploaded.

\textbf{Fusion:} If \ac{BN} is used for normalization, we fuse the convolution operation with the following \ac{BN} operation~\citep{ioffe2015batch, jacob2017quantization} in frozen layers (we study the application of other normalization techniques in \cref{sec:experiments}).
\Iac{BN} layer normalizes each channel to zero mean and unit variance followed by a trainable scale~$\gamma_i$ and bias~$\beta_i$.
The statistics of frozen blocks that do not require intermediate gradients (type (a)) stay constant over time. Hence, we can express the \ac{BN} operation as a linear operator with~$y_{i_{\text{BN}}}$ being the channel-wise output of the \ac{BN} operation, $y_{i_{\text{CL}}}$ the \ac{CL}'s output, and~$\epsilon$ a small number for stability
\begin{equation}
\label{eq:fusion}
\textstyle
y_{i_{\text{BN}}}
    = \frac{\gamma_i} {\sqrt{\sigma_i^2 + \epsilon}}\cdot y_{i_\text{CL}}
    + (\frac{ - \mu_i \gamma_i}{\sqrt{\sigma_i^2 + \epsilon}} + \beta_i),
\end{equation}
where the coefficient of~$y_{i_\text{CL}}$ is a new combined scale, referred to as~$\hat \gamma_i$, and the second summation term is a new combined bias, referred to as~$\hat \beta_i$. To fuse the \ac{BN} operation with the preceding \ac{CL}, we express the \ac{CL} as~$y_{i_\text{CL}} = \bm W_i{\cdot}x$, and plug the output~$y_{i_\text{CL}}$ into the \ac{BN} operator: $y_{i_\text{BN}} = (\hat \gamma_i \bm W_i) {\cdot} x +  \hat \beta_i$. This gives a scaled version of the original kernel~$\hat{ \bm W}_i = \hat \gamma_i \bm W_i$ with a new bias~$\hat \beta_i$. The same can be applied for type (c) layers, with the only difference being that $\mu$ and $\sigma^2$ values are only valid for a limited number of mini-batches.
\emph{In summary, the forward pass of type~(a) and type~(c) blocks is simplified by fusing three operators, reducing the number of operations.}

\textbf{Quantization:} Quantization of operators in \acp{NN} is usually used for inference.
In contrast, we apply the idea for training; however, we \textbf{quantize only parts of the \ac{NN} that are frozen.} Note that this is different from \emph{quantization aware training}, since all trained parameters remain in full precision. Therefore, type~(a) and~(c) blocks are quantized, i.e., the fused convolution is performed in \texttt{int8} instead of \texttt{float32}. 
Additionally, in type~(c) blocks, we also quantize the calculation of the intermediate gradients in the backward pass, i.e., the fused transposed convolution.
\emph{Consequently, the remaining operations in the forward and backward pass of frozen blocks require less time for execution and have a lower memory footprint.}

Quantization of operations introduces quantization noise in the forward pass and the backward pass of frozen layers, thereby affecting the training. Additionally, updating the fused layers' statistics only at the beginning of the round introduces an error. We demonstrate in our experiments that the benefits of increased efficiency w.r.t. training time and memory footprint outweigh the added noise. We quantify the benefits and the effects of quantization noise in detail in an ablation study in~\cref{sec:ablation_study}.
\emph{In summary, freezing, fusion, and quantization of selected blocks lower the computational complexity, while still allowing less capable devices to calculate parameter gradients of other blocks in full precision based on the full \ac{NN}. This novel combination has not yet been exploited for training.}

\subsection{Implementation in PyTorch}
\label{subsec:implementation_in_pytorch}
We implement the presented training scheme in PyTorch 1.10~\citep{PyTorch}, which supports \texttt{int8} quantization. While the following description could also be applied to other quantization schemes (e.g., \texttt{int4}, \texttt{float16}), as of now, PyTorch only provides the necessary operators in the backends \texttt{FBGEMM} and \texttt{QNNPACK} for \texttt{int8}. Quantization levels, such as \texttt{int4}, could further lower the training time but at the same time could also have an impact on the accuracy. Contrary to quantization for inference, the layers' input scales, as well as the BN layers' statistics change throughout the training. Our implementation enables real-world training time and memory reduction through a combination of on-the-fly scale calculation and statistics from the server.

\textbf{Quantization in the Forward Pass:} To preserve a high accuracy, the quantization of a PyTorch tensor~$x$ requires a scale~$s_x$ to optimally utilize the \texttt{int8} range. We calculate the scale by using
\begin{align}
\label{eq:scale}
    s_x = s(x) = 2\cdot\text{max}(|\text{max}(x)|,| \text{min}(x)|)/127.0.
\end{align}
Quantized operators (e.g., \texttt{linear}, \texttt{conv2d}, and \texttt{add}) take a quantized tensor~$\bar x$ (and quantized weights) as input. In the used backends calculations are done in \texttt{int8} arithmetic, but the accumulation of the result is done in \texttt{int16/32}. Therefore, each quantized operator requires an output scale~$s_o$ to map the \texttt{int16/32} output to \texttt{int8}. This output scale~$s_o$ is calculated using~\cref{eq:scale}. For blocks of type (a), for every mini-batch, we calculate the scale of the input tensor $s_x$ on the fly at the beginning of the first type (a) block. The input gets quantized and stays in the quantized representation throughout its forward propagation through type (a) blocks. For input scales~$s_x$, the scale calculation results in negligible overhead, as~$x$ is already available in its \texttt{float32} representation. However, the output scales~$s_o$ depend on the output of an operation (\texttt{linear}, \texttt{conv2d}, \texttt{add}), therefore, can not be calculated a priori without performing the full operation in float. Because of this PyTorch limitation, the output scale $s_o$ and the \ac{BN} layer's statistics ($\sigma^2$ and~$\mu$ from \cref{eq:fusion}) are obtained from the server and are only set once per \ac{FL} round.

\textbf{Quantization in the Backward Pass:} Out of the box, PyTorch's \emph{Autograd} system does not support a quantized backward pass but expects \texttt{float32} values for each calculated gradient. We implement a custom PyTorch \emph{Module} for blocks of type (c) based on a custom \emph{Autograd Function} that encapsulates all quantized operations in one \emph{backward} call. Due to this limitation, a quantization/dequantization is required, for each block each mini-batch. The intermediate gradients' scale $s_g$ is calculated on the fly. These limitations are inherently considered in our experimental results (profiling), i.e., fixing these limitations of PyTorch would further increase the efficacy of CoCoFL. These overheads are minor compared to the speedups gained through quantization, since large convolution operations dominate the training time (e.g., we measure a 6\% overhead of scale/quantization/dequantization of type (c) blocks in MobileNet~\citep{howard2017mobilenets} but gain a reduced computation time by a factor of 1.3). Similar to type (a) blocks, the output scales of operators (e.g., transposed convolution) in the backward pass are obtained from the server.

Type (b) blocks (trained blocks) require only minor modifications. In order to acquire the scale of the operations, PyTorch forward and backward \texttt{hooks} are used to calculate~$s_o$. For trained blocks, $s_o$ can be efficiently acquired since trained blocks' operators calculate regular \texttt{float32} outputs. Together with the trained parameters, these scales are uploaded to the server and averaged alongside the \ac{NN} parameters. The scales only have to be calculated in the last mini-batch of a training round and result in negligible overhead. To perform fusion, as presented in \cref{sec:technique}, devices that train a respective block upload their \ac{BN} statistics. The statistics are averaged alongside the parameters and distributed to the devices that require the respective statistics for fusion.

\textbf{Transformer-Specific Implementation Details:} We treat \emph{encoder} layers as blocks, and quantize linear, layernorm, and ReLU operations for type (a) blocks. Due to PyTorch limitations, the attention mechanism has to remain in \texttt{float32}. In type (c) blocks, linear layers and their intermediate gradient calculations get quantized. Further details are provided in~\cref{sec:appdx_details}.

\section{Overall CoCoFL Algorithm}
\label{sec:technique_block_selection}
Partial freezing and quantization enables to adjust the required communication, computation, and memory resources by selecting which blocks to train or freeze.
We present CoCoFL, which enables each device to select the trained/frozen blocks based on its available resources, and the required changes in aggregation at the server, in order to maximize the accuracy under constrained resources.

\textbf{Heuristic Configuration Selection}:
A selection of trained blocks is a training \emph{configuration}~$A\in\mathcal A$.
The set of all configurations of \iac{NN} with $N$~blocks comprises $|\mathcal A| = 2^N$ configurations (each block is either trained or frozen). In each round, each device can select a separate configuration. Therefore, the total search space in each round is~$2^{N{\cdot} |\mathcal C|}$, which is infeasible to explore in its entirety, and impractical as it depends on the parameters like the \ac{NN} structure. Simplifying the search space by assigning a separate quality measure to each configuration also does not work, since the accuracy after training with a certain configuration depends also on the configurations used by other devices. Therefore, heuristic optimization is required. Simple deterministic heuristics like selecting configurations that train the maximum number of blocks at once show bad performance, as some blocks would never get trained. As another example, a round-robin selection would lead to all devices selecting the same configuration, where our observations have shown that this leads to lower accuracy (we provide experimental results in~\cref{appdx:block_selection_ablation_study}).
Therefore, CoCoFL selects a random configuration on each device based on its available resources. Thereby, the probability that many devices select the same configuration is negligible, while eventually all blocks within a device's capability get trained. An additional benefit of this scheme is that no signaling between the server and devices to transmit the available resources and selected configurations is required, which otherwise could slow down the overall \ac{FL} process and prevent scalability.

\begin{algorithm*}[t]
	\caption{Each Selected Device~$c$ (Client) in Each Round}
	\footnotesize
	\label{algo:device}
	\begin{algorithmic}
	    \Require $s, t_c, m_c, \mathcal D_c, T, S_c, M_c$ \Comment{profiling information of the device, data, available resources}
		\State receive $w^{(r)}$ from the server \Comment{initial parameters at the beginning of the round}
		\State $\mathcal A_f \gets \{ A \in \hat{\mathcal A} : t_c(A) \le T\ \land\ m_c(A) \le M_c\ \land\ s(A) \le S_c \}$ \Comment{feasible configurations (Eq.~\ref{eq:constraint})}
		\State $\mathcal A_{\text{max}} \gets \{ A_j \in \mathcal A_f: \forall A_k\in \mathcal A_f : A_j \not\subset A_k  \}$ \Comment{discard non-maximal configurations}
		\State $A_c^{(r)} \gets \text{random choice}(\mathcal A_{\text{max}})$ \Comment{select random configuration}
		\State $(w_\text{train}, w_\text{quant}) \gets \text{apply}(A_c^{(r)}, w^{(r)})$ \Comment{apply the configuration (freeze/quantize blocks)}
        \State{$\tilde{w}_\text{train} \gets$ train model $(w_\text{train},w_\text{quant})$ with local data $\mathcal D_c$} \Comment{local training}
		\State{send $\tilde{w}_\text{train}$ to the server} \Comment{parameter update}
	\end{algorithmic}
\end{algorithm*}

\begin{algorithm*}[t]
	\caption{FL Server (Synchronization and Aggregation)}
	\footnotesize
	\label{algo:server}
	\begin{algorithmic}
		\State $w^{(1)} \gets$ random initialization
		\For{\textbf{each} round~$r=1,2,\ldots,R$}
		    \State $\mathcal C^{(r)}\gets$select devices
			\State broadcast $w^{(r)}$ to selected devices~$\mathcal C^{(r)}$
			\State \textbf{for each} $c\in\mathcal C^{(r)}$ \textbf{do} receive $\tilde{w}_{\text{train},c}$ from device $c$
			\For{\textbf{each} block~$i$} \Comment{{aggregation}}
    			\State $\mathcal{C}_i \gets \{ c : \tilde{w}_{\text{train},c} \text{ contains block } i \}$ \Comment{{devices that have trained block~$i$}}
    			\State $w^{(r+1)}_i \gets
    			    \left(1-\frac{|\mathcal{C}_i|}{|\mathcal C^{(r)}|}\right) \cdot w^{(r)}_i
    			    +
    			    \frac{1}{|\mathcal C^{(r)}|} \sum_{c \in \mathcal{C}_i} \tilde{w}_{\text{train},c}(i)
    			$
    		\EndFor
		\EndFor
	\end{algorithmic}
\end{algorithm*}

To be able to select configurations w.r.t.~the available resources, we need to quantify the resource requirements per configuration.
We obtain this information through design-time profiling of a real implementation of our presented freezing and quantization scheme on real devices, but an analytical model of the resources could also be employed.
Profiling takes several seconds per configuration.
For instance, profiling MobileNet takes \SI{4.7}{\second} on average on the {x64} target platform.
It is, therefore, infeasible to profile all $2^N$ configurations per device, which would take several months.
We solve this by only considering configurations~$\hat{\mathcal{A}}\subseteq\mathcal A$ that train a single contiguous range of blocks.
This reduces the search space to~$|\hat{\mathcal{A}}| = \frac{N(N+1)}{2}$, i.e., \SI{17}{\minute} for MobileNet on {x64}.
If resources can be estimated much faster, relaxing this restriction could further improve our technique.
The run-time algorithm for devices is outlined in~\cref{algo:device}.
At the beginning of each round, each device determines the set of feasible configurations~$\mathcal{A}_f{\subseteq}\hat{\mathcal{A}}$, given its currently available resources.
We then discard all configurations that train a subset of blocks trained by another feasible configuration, i.e., we only keep maximal configurations~$\mathcal{A}_{\text{max}}{\subseteq}\mathcal{A}_f$, thereby maximizing the accuracy by fully exploiting the available resources.
Each device selects a random remaining configuration, which results in different configurations being trained on different devices without requiring any synchronization between devices.
Finally, the selected fusion and quantization configuration is applied, and the \ac{NN} is trained.

\textbf{Aggregation of Partial Updates}:
Each device~$c$ only uploads updates of the blocks that were trained in full precision~($\tilde{w}_{\text{train},c}$), hence, not frozen or quantized. The server (\cref{algo:server}) weighs the updates based on the number of devices that have trained each block to account for partial training on the devices.
\cref{fig:system_model} shows CoCoFL in a nutshell.

\section{Experimental Evaluation}
\label{sec:experiments}
Partial quantization of \ac{NN} models results in hardware-specific gains in execution time and memory. Hence, our evaluation follows a hybrid approach, where we profile on-device training loops on real hardware and take the profiling information to perform simulations of distributed systems. This allows for the evaluation of large systems with hundreds or thousands of devices.

\begin{figure*}
    \centering
    \trimbox{0.1cm 0.4cm 0.1cm 0.1cm}{
        \includegraphics[page=1]{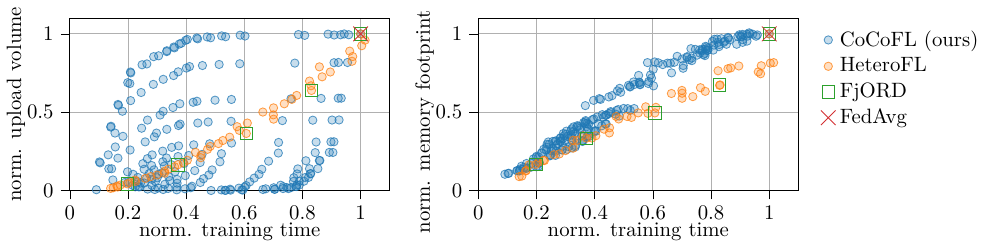}
    }
    \caption{Profiling of MobileNet regarding upload volume, training time, and memory footprint, normalized to training the full \ac{NN} with FedAvg. Each marker represents a training configuration. The figure shows that CoCoFL supports the same range w.r.t. to the constraints as HeteroFL and FjORD but enables independent adjustability of computation/memory and communication.}
    \label{fig:profiling}
\end{figure*}

\subsection{Evaluation Setup}
\label{subsec:evaluation_setup}
\textbf{Profiling Setup and Results:}
We employ two different hardware platforms to factor out potential micro\-archi\-tecture-dependent peculiarities w.r.t.\ quantization or freezing: x64 AMD Ryzen 7 and a Raspberry Pi with an ARMv8 CPU. For each configuration in~$\hat{\mathcal{A}}$, we measure the execution time, maximum memory usage, and upload volume. The measurements are stored in a lookup table for the \ac{FL} simulations. \Ac{NN}-specific details about~$N$, $|\hat{\mathcal{A}}|$, and the used platform are given in~\cref{tab:hyperparameters}, further details in~\cref{sec:appdx_profiling}.
For simplicity in the implementation, we select one \emph{skip connection block} in ResNet, MobileNet, and DenseNet as the smallest entity that is either trained or frozen. This choice allows for limited implementation overhead, as the structure is repeatedly used in the \acp{NN}. The block granularity could be further increased by selective training and freezing of \ac{CL} layers within a skip connection block, but it would require more individual cases to be implemented. Exemplary profiling results of MobileNet on x64 are shown in \cref{fig:profiling}, where all quantities are normalized to FedAvg (full training of the \ac{NN}). Using the same setup, we profile training with subsets of the \ac{NN}'s filters, as employed by HeteroFL and FjORD. In CoCoFL, a configuration refers to a specific selection of blocks that are frozen, quantized, and fused. In HeteroFL/FjORD, a configuration refers to a specific ratio of filters that are trained and filters that are dropped. As a consequence, the trade-offs vary. Our results show that the combination of freezing, fusion, and quantization allows training with configurations that reduce the execution time by up to 90\% and the memory footprint by 89\% compared to full training of the \ac{NN}, a similar range as HeteroFL and FjORD. Further, CoCoFL enables independent adjustability of computation/memory and communication, giving us a higher degree of freedom to select a configuration that utilizes the resources at the device, therefore, more efficiently utilizing available resources. Contrary to that, training subsets results in a tightly coupled reduction of resources. However, in some cases, CoCoFL is required to pick a non-optimal configuration w.r.t. computation, to satisfy a memory constraint.
\begin{table*}
\caption{Hyperparameters of \ac{FL} experiments.}
\centering
\begin{adjustbox}{width=\columnwidth,center}
\small{
\renewcommand{\arraystretch}{0.6}
	\begin{tabular}{@{}lp{1.5cm}p{1.2cm}p{1.2cm}p{1.4cm}p{1.4cm}p{1.2cm}p{2cm}@{}}
		\toprule
		Hyperparameters & \footnotesize{DenseNet \mbox{CIFAR10}/ (CINIC10)} &  \footnotesize{MobileNet \mbox{CIFAR10}} & \footnotesize{MobileNet \mbox{CIFAR10} \mbox{GroupNorm}} & \footnotesize{ResNet50 CIFAR100} & \footnotesize{ResNet18 FEMNIST} & \footnotesize{MobileNet \mbox{XChest}} & \footnotesize{Transformer \mbox{IMDB}/ (Shakespeare)} \\
		\midrule
		Rounds~$R$ & $800$ & $1000$ & $1000$ & $800$ & $600$ & $1000$ & $1000$ \\
		Amt. data~$|\mathcal{D}|$ & $50$K ($90$K)& $50$K & $50$K & $50$K & $640.5$K & $12.7$K & $40$K ($200$K) \\
		$|\mathcal{C}^{(r)}|/|\mathcal{C}|$ & $10/100$ & $10/100$ & $10/100$ & $10/100$ & $35/3500$ & $10/100$ & $10/100$ \\
		$\eta$-decay ($\times0.1$) & $[750]$ & $[600,800]$ & $[800]$ &  $[750]$ & $[400]$ & $[600]$ & $[800]$ \\
		Weight Decay & $0.001$ & $0.01$ & $0.001$ & $0.01$ & $0.01$ & $0.01$ & - \\
		Nb. configs~$|\hat{\mathcal{A}}|$ & 253 & 210 & 210 & 171 & 55 & 210 & 35 \\
		Nb. blocks~$N$ & 23 & 21 & 21 & 19 & 11 & 21 & 8\\
		Platform & ARM & x64 & x64 & x64 & ARM & x64& x64 \\
		\bottomrule
	\end{tabular}
 }
 \end{adjustbox}
	\label{tab:hyperparameters}
\end{table*}

\textbf{\ac{FL} Setup and Hyperparameters:} We evaluate our technique in \iac{FL} system, using the profiling results. For each experiment, we distribute the data from the datasets CIFAR10/100~\citep{krizhevsky2009learning}, FEMNIST~\citep{cohen2017emnist}, CINIC10~\citep{darlow2018cinic}, XChest~\citep{Wang_2017_CVPR}, IMDB~\citep{imdb}, and Shakespeare~\citep{caldas1812leaf} to devices in~$\mathcal C$. We evaluate ResNet~\citep{gao2020residual}, DenseNet~\citep{densenet}, MobileNet~\citep{howard2017mobilenets}, and Transformer~\citep{vaswani2017attention} \ac{NN} models. In each round, a subset $\mathcal{C}^{(r)}$ is selected for participation. Devices are randomly grouped in three equally sized sets. The set of \emph{strong} devices is capable of training the full \ac{NN} and uploading all parameters, with no memory constraints. The round time $T$ is set to the time a \emph{strong} device requires to finish one training round. The set of \emph{medium} devices has~$2/3$ of the computational and memory resources of the \emph{strong} devices. Hence, to match the round time~$T$, the set of \emph{medium} devices has to select configurations that reduce required computations to 2/3 of \emph{strong} devices. The set of \emph{weak} devices has $1/3$ of the computation and memory capabilities of the \emph{strong} devices. We model the communication budgets of \emph{medium} and \emph{weak} devices randomly over rounds to simulate an environment with varying communication channel quality, s.t.~$ S_{c_{\text{medium, weak}}}^{(r)} \sim \mathcal{U} (\frac{S_{\text{strong}}}{2}, S_{\text{strong}})$. We compare CoCoFL to several baselines: state-of-the-art HeteroFL and FjORD, which are the closest to our technique, as both allow for a per-device reduction of computational resources, as well as upload volume, and memory. Additionally, we compare to a theoretical bound and a straightforward baseline that drops all but the \emph{strong} devices from \ac{FL} training.
\begin{itemize} [noitemsep, nolistsep]
    \item Centralized: All data is centralized (on one device), serves as a theoretical upper bound.
    \item FedAvg (full resources)~\citep{mcmahan2017communicationefficient}: \ac{FL} is applied, but all devices have full (homogeneous) resources, hence, FedAvg has only one configuration, that is training the full network. This baseline serves as a theoretical upper bound.
    \item HeteroFL~\citep{diao2020heterofl}: A FedAvg variant that drops a number of the \ac{CL} filters, defined by a shrinkage ratio, where each ratio represents a training configuration. We set it to the maximum that a device can~train.
    \item FjORD~\citep{horvath2021fjord}: Similarly, state-of-the-art FjORD drops \ac{CL} filters to reduce resources. In difference to HeteroFL, each device trains different drop levels (within their capabilities). Devices switch each mini-batch between a feasible configuration. The paper proposes to use $[20,40,60,80,100]\%$ of the filters, which results in 5 configurations.
    \item FedAvg~\citep{mcmahan2017communicationefficient}: Devices that can not train the \ac{NN} (e.g., due to limited memory) are dropped from the training (therefore also their data). The reduced set of devices performs FedAvg. This serves as a naive baseline and is known to be used in production use cases~\citep{yang2018applied}. 
\end{itemize}

We train with the optimizer \ac{SGD} with an initial learning rate~$\eta$ of $0.1$. For a fair comparison we do not use momentum, as FjORD is incompatible with a stateful optimizer. The remaining \ac{NN}-specific hyperparameters, learning rate decay, and weight decay are given in~\cref{tab:hyperparameters}. For each \ac{FL} experiment, we report the average accuracy and standard deviation after~$R$ rounds of training using $3$ independent seeds. We study several data split scenarios: First, \iac{IID} case, where data is randomly distributed to all devices, and hence, every device has about the same number of samples per class. Second, we study a non-\ac{IID} case, where we vary the non-\ac{IID}-ness with the value of~$\alpha$ of a Dirichlet distribution, similar to~\cite{hsu2019measuring}. Hereby, the number of samples per class varies between devices. Thirdly, we consider a scenario where data is \emph{resource correlated non-\ac{IID}} (\emph{rc-non-\ac{IID}}). This means that information about certain classes is only available on specific device groups, increasing the necessity to include them in the \ac{FL} process (\cref{fig:data_distribution}). This has recently been identified as a relevant use case in real-world deployments~\citep{maeng2022towards}. Similarly, the rate of rc-non-\ac{IID}-ness is controlled with~$\alpha$.

\begin{table*}
\centering
\caption{Accuracy (Top 1) in \% for DenseNet, MobileNet, ResNet18, ResNet50, and Transformer. For XChest the F1 macro score is given (unbalanced data). In almost all scenarios CoCoFL outperforms the baselines, reaching higher final accuracies.}
\label{tab:acc}
\color{black}
\begin{adjustbox}{width=\columnwidth,center}
\small{
\begin{tabular}{@{}l@{\hskip + 5pt}c@{\hskip + 8pt}c@{\hskip + 8pt}c@{\hskip + 8pt}c@{\hskip + 8pt}c@{\hskip + 8pt}c@{\hskip + 8pt}c@{\hskip + 8pt}c@{\hskip + 8pt}c@{\hskip + 8pt}c@{\hskip + 8pt}c@{\hskip + 8pt}}
\toprule
Topology & \multicolumn{3}{c}{\textbf{DenseNet}} &\multicolumn{5}{c}{\textbf{MobileNet}} & \multicolumn{2}
{c}{\textbf{ResNet18}}\\ \cmidrule(l{1pt}r{5pt}){2-4} \cmidrule(l{1pt}r{5pt}){5-9}
\cmidrule(l{1pt}r{5pt}){10-11}
Setting & \multicolumn{3}{c}{\textbf{CIFAR10}}& \multicolumn{2}{c}{\textbf{CIFAR10}} &\multicolumn{3}{c}{\textbf{CIFAR10 (w. GroupNorm)}} & \multicolumn{2}
{c}{\textbf{FEMNIST}}\\ \cmidrule(l{1pt}r{5pt}){2-4} 
 \cmidrule(l{1pt}r{5pt}){5-6} 
 \cmidrule(l{1pt}r{5pt}){7-9}
\cmidrule(l{1pt}r{5pt}){10-11}
Dirichlet $\alpha$ & $-$ (iid) & n.-iid@0.1 & rc@0.1 & $-$ (iid) & rc@0.1 & - (iid) & n.-iid@0.1& rc@0.1 & $-$ (iid) & rc@0.1\\
\midrule
Centralized & \multicolumn{3}{c}{87.8$\pm$0.2} &\multicolumn{2}{c}{87.0$\pm$0.7} &\multicolumn{3}{c}{82.5$\pm$1.1}& \multicolumn{2}{c}{88.1$\pm$0.0}\\
FedAvg (f. res.)& 84.3$\pm$0.1 & 75.6$\pm$1.5 &  74.9$\pm$3.3  & 84.9$\pm$0.2 & 77.4$\pm$2.7 & 
76.1 $\pm$ 0.1 & 69.4 $\pm$ 0.5 & 72.9 $\pm$ 1.4 &
86.2$\pm$0.1 & 82.9$\pm$1.2\\
\midrule
CoCoFL (ours) &\textbf{82.0}$\pm$\bftab{0.2} &  \bftab{71.9}$\pm$\bftab{1.8} & \textbf{68.8}$\pm$\textbf{4.6} & \textbf{83.2}$\pm$\textbf{0.3} & \textbf{72.4}$\pm$\textbf{2.9} & 
\textbf{71.3} $\pm$ \textbf{0.1}&  \textbf{61.2} $\pm$ \textbf{1.4} & \textbf{63.6}  $\pm$ \textbf{3.5} & 
85.0$\pm$0.1 & \textbf{81.5}$\pm$\textbf{0.6}\\

FjORD & 73.7$\pm$0.1 &60.4$\pm$2.1 & 48.8$\pm$6.8 & 79.1$\pm$0.3 & 51.9$\pm$7.3 &
64.4 $\pm$ 0.6 & 42.9 $\pm$ 1.7 & 47.0 $\pm$ 5.0 &
85.5$\pm$0.0 & 69.3$\pm$8.3\\
HeteroFL & 76.4$\pm$0.3 &  64.0$\pm$2.4 & 51.2$\pm$7.4 &79.5$\pm$0.2 & 53.0$\pm$7.6 &
64.8 $\pm$ 0.1 & 55.2 $\pm$ 0.3 & 47.5 $\pm$ 5.0 &
85.9$\pm$0.0 & 70.9$\pm$5.8\\
FedAvg& 76.5$\pm$0.1 & 60.4$\pm$4.2 & 50.9$\pm$7.5 & 78.1$\pm$0.4 & 49.9$\pm$8.7 &
56.2 $\pm$ 0.8 & 52.8 $\pm$ 1.1 & 45.9 $\pm$ 4.7&
\textbf{86.1}$\pm$\textbf{0.1} & 64.9$\pm$7.8\\ 
\end{tabular}
}
\end{adjustbox}
\begin{adjustbox}{width=\columnwidth,center}

\small{
\begin{tabular}{@{}l@{\hskip + 5pt}c@{\hskip + 5pt}c@{\hskip + 5pt}c@{\hskip + 5pt}c@{\hskip + 5pt}c@{\hskip + 5pt}c@{\hskip + 5pt}c@{\hskip + 5pt}c@{\hskip + 5pt}c@{\hskip + 5pt}c@{\hskip + 5pt}c@{\hskip + 5pt}}
\toprule
Topology & \multicolumn{2}{c}{\textbf{ResNet50}} & \multicolumn{3}{c}{\textbf{DenseNet}} &\multicolumn{3}{c}{\textbf{MobileNet(large)}} & \multicolumn{1}{c}{\textbf{TF}} & \multicolumn{2}{c}{\textbf{TF-S2S}}\\
\cmidrule(l{5pt}r{5pt}){2-3}
\cmidrule(l{5pt}r{5pt}){4-6}
\cmidrule(l{5pt}r{5pt}){7-9}
\cmidrule(l{10pt}r{10pt}){10-10}
\cmidrule(l{10pt}r{10pt}){11-12}

Setting & \multicolumn{2}{c}{\textbf{CIFAR100}} & \multicolumn{3}{c}{\textbf{CINIC10}} &\multicolumn{3}{c}{\textbf{XChest}} & \multicolumn{1}{c}{\textbf{IMDB}} & \multicolumn{2}{c}{\textbf{Shakespeare}}\\
\cmidrule(l{5pt}r{5pt}){2-3}
\cmidrule(l{5pt}r{5pt}){4-6}
\cmidrule(l{5pt}r{5pt}){7-9}
\cmidrule(l{10pt}r{10pt}){10-10}
\cmidrule(l{10pt}r{10pt}){11-12}

Dirichlet $\alpha$ & $-$ (iid) & rc@0.1 & $-$ (iid) & n.-iid@0.1 & rc@0.1 & $-$ (iid) & n.-iid@0.5 & rc@0.5 & $-$ (iid) & $-$ (iid) & $-$rc (Leaf)\\
\midrule
Centralized & \multicolumn{2}{c}{61.6$\pm$0.4} & \multicolumn{3}{c}{80.5$\pm$0.2} & \multicolumn{3}{c}{94.2$\pm$0.2} & 84.7$\pm$0.7 & \multicolumn{2}{c}{52.9$\pm$0.7}\\
FedAvg (f. res.)& 57.0$\pm$0.3 & 53.0$\pm$0.6 & 77.2$\pm$0.1 & 53.9$\pm$2.3 & 65.1$\pm$1.1 & 94.1$\pm$0.3  & 85.9$\pm$1.8 & 93.2$\pm$0.2 & 82.6$\pm$0.4 & 49.1$\pm$0.1 & 49.4$\pm$0.1\\
\midrule
CoCoFL (ours) & \textbf{52.5}$\pm$\textbf{0.2} & \textbf{41.8}$\pm$\textbf{2.5} & \textbf{73.6}$\pm$\textbf{0.1} & \textbf{53.5$\pm$4.3} & \textbf{52.4}$\pm$\textbf{7.3} & \textbf{91.3}$\pm$\textbf{0.3} & \textbf{73.0}$\pm$\textbf{6.4} & \textbf{87.3}$\pm$\textbf{3.8} & \textbf{82.5}$\pm$\textbf{0.5}& \textbf{49.3}$\pm$\textbf{0.3} & \textbf{49.1}$\pm$\textbf{0.3}\\
FjORD & 43.6$\pm$0.8 & 29.6$\pm$4.3 & 65.1$\pm$0.7 & 49.2$\pm$2.2 & 41.1$\pm$6.9 & 66.3$\pm$0.9 & 52.7$\pm$3.9 & 62.4$\pm$0.8 & \phantom{\textsuperscript{\textbf{1}}}78.5$\pm$0.7\textsuperscript{\textbf{1}} & 42.9$\pm$0.5 & 43.0$\pm$0.3 \\
HeteroFL & 45.9$\pm$0.7 & 31.0$\pm$2.3 &  69.4$\pm$0.2 & 50.4$\pm$2.4 & 43.4$\pm$7.2 & 69.4$\pm$1.0 & 65.0$\pm$0.9 & 65.4$\pm$1.6  & \phantom{\textsuperscript{\textbf{1}}}79.2$\pm$0.3\textsuperscript{\textbf{1}} & 44.1$\pm$0.2 & 44.1$\pm$0.2\\
FedAvg& 35.2$\pm$0.2 & 23.7$\pm$0.4 & 67.7$\pm$0.4 & 48.3$\pm$2.5 & 42.4$\pm$7.2 & 68.2$\pm$1.0 & 67.0$\pm$0.6 & 66.8$\pm$1.4 & 78.5$\pm$0.6 & 40.5$\pm$0.3 & 40.3$\pm$0.1\\ 
\bottomrule
\end{tabular}
}
\end{adjustbox}
\begin{tablenotes}
 \item \small{\textsuperscript{\textbf{1}}No configuration for \emph{weak} devices available, therefore \emph{weak} devices are dropped.}
\end{tablenotes}
\end{table*}

\subsection{FL Results}
\cref{tab:acc} presents the accuracy results given in \%. For XChest (unbalanced) the F1 macro score is given.

\textbf{Vision Models:} For \ac{IID} data, CoCoFL performs close to FedAvg (full res.), improving the final accuracy over the baselines by $\SI{5.5}{\pp}$ for DenseNet (CIFAR10), by $\SI{3.7}{\pp}$ for MobileNet (CIFAR10), and by $\SI{6.6}{\pp}$ for ResNet50 (CIFAR100). Similar trends can be seen for CINIC10. This clearly indicates that CoCoFL uses the available resources on devices more effectively. An outlier is the FEMNIST dataset, where FedAvg reaches the highest accuracy despite dropping 2/3 of the devices. We attribute this to the high number of redundant samples in the dataset ($10$K per class compared to with $5$K and $500$). Contrary to that, if the number of samples is more limited, as it is the case in the XChest experiments ($12$K samples total), the advantage of CoCoFL over the baselines increases.

The necessity to include less capable devices in the \ac{FL} training is more clearly visible in cases with rc-non-\ac{IID}. In \cref{tab:acc}, it can be seen that $\alpha=0.1$ results in a larger gap between the upper bound and the naive baseline, demonstrating the importance of involving all devices in the training. CoCoFL enables \emph{weak} and \emph{medium} devices to contribute to the global model, reaching up to $\SI{20}{\pp}$ higher accuracy compared with the state of the art. The reason is that CoCoFL allows \emph{weak} and \emph{medium} devices to calculate gradients based on the full \ac{NN}. In the case of DenseNet and MobileNet, FjORD and HeteroFL even perform close or inferior to the naive baseline, which excludes \emph{weak} and \emph{medium} devices from the training, failing to preserve fairness (accuracy parity) among devices. Similar conclusions can be driven from ResNet18/50, albeit FjORD and HeteroFL perform a bit better in these settings compared with FedAvg. For XChest we present results with $\alpha=0.5$ since we observe that $0.1$ leads to a complete separation of the binary labels, causing all algorithms to fail to learn at all.

\begin{figure}
    \begin{subfigure}[t]{0.5\textwidth}
        \trimbox{0.0cm 0.0cm 0.0cm 0.0cm}{
            \includegraphics[page=1]{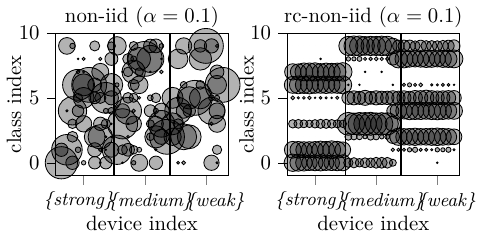}
        }
        \caption{Demonstration of non-iid data (left) and resource-correlated-non-iid data (right). The size of the circles represents the number of samples.}
        \label{fig:data_distribution}
    \end{subfigure}
    \begin{subfigure}[t]{0.5\textwidth}
     \trimbox{0.0cm 0.0cm 0.0cm 0.0cm}{
        \includegraphics[page=1]{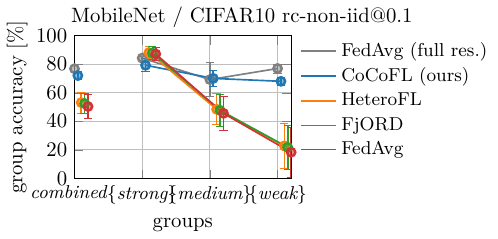}
        }
         \caption{Group accuracies of MobileNet with CIFAR10 and rc-non-iid@0.1 data. CoCoFL preserves fairness among groups, since \emph{weak} and \emph{medium} devices achieve similar group accuracies as if they would have full resources.}
         \label{fig:fairness}
    \end{subfigure}
    \caption{\cref{fig:data_distribution} visualizes resource-correlated non-iid, while \cref{fig:fairness} shows the effects of this distribution w.r.t. the fairness (group accuracy) in CoCoFL and baselines.}
\end{figure}

\textbf{Fairness in Rc-non-iid Scenarios:} To quantify the contribution \emph{medium} and \emph{weak} devices make in the training, we calculate the device-specific accuracy per group (\emph{group} accuracy), where the class accuracies are weighted by the groups' class densities. As it can be seen for MobileNet (rc-non-iid@0.1) in~\cref{fig:fairness}, CoCoFL achieves group accuracies of 79\%/70\%/68\% for \emph{strong}, \emph{medium}, and \emph{weak} devices, hence, close to the accuracy parity of FedAvg (with full resources). The baselines HeteroFL and FjORD reach 88\%/48\%/23\%, meaning less capable devices can not make a meaningful contribution to the global model, hence, lowering the fairness among the device groups.

\textbf{Other Normalization Techniques}: We replace \ac{BN} with GroupNorm~\citep{wu2018group} in MobileNet to test the robustness of CoCoFL w.r.t. other normalization techniques in vision tasks. For MobileNet with CIFAR10, we observe that the overall accuracy is lower in all evaluated algorithms (\cref{tab:acc}). However, the general trends are similar to MobileNet with \ac{BN}, i.e., independent from the normalization, CoCoFL outperforms the state of the art. Additionally, we evaluate the rc-non-iid scenario, where CoCoFL reaches group accuracies of $63.7\pm3.5\%$, $60.7\pm5.6\%$, and $58.3 \pm9.8\%$ for \emph{strong}, \emph{medium}, and \emph{weak} devices, whereas FjORD and HeteroFL reach $80.7\pm7.2\%$/$42.8\pm8.5\%$/$18.3\pm14.4\%$ and $81.3\pm7.1\%$/$42.8\pm9.0\%$/$19.0\pm13.5\%$, respectively. Thus, CoCoFL provides better fairness independent of the normalization technique.

\textbf{NLP Models}: To show the applicability of CoCoFL to \ac{NLP} problems, we adapt our freezing and quantization scheme (\cref{sec:technique}) for Transformer. We study text classification with the IMDB dataset and next character prediction with the Shakespeare dataset. The Transformer model uses~6 \emph{encoder} layers, with embedding size of~128, hidden size of~128, 2 attention heads, and a single linear decoder layer. For IMDB the sequence length is~512, for Shakespeare~80. For Shakespeare rc-non-iid, we follow the non-iid scheme from Leaf~\citep{caldas1812leaf}, such that different plays (total of~25) are distributed over different device groups. The results in~\cref{tab:acc} show that CoCoFL reaches significantly higher final accuracies than the state of the art.

\emph{In summary, CoCoFL reaches higher accuracies in almost all presented scenarios. Additionally, CoCoFL preserves fairness (accuracy parity) by enabling constrained devices to contribute to the global model. We attribute this large accuracy gap w.r.t the baselines to the fact that CoCoFL allows any device to calculate gradients based on the full \ac{NN}, while still reducing required resources, as opposed to state-of-the-art techniques that calculate gradients on subsets of the filters of the \ac{NN}.}

\subsection{Ablation Study}
\label{sec:ablation_study}
\begin{figure}
  \begin{subfigure}[t]{0.48\textwidth}
        \trimbox{0.0cm 0.0cm 0.0cm 0.0cm}{
            \includegraphics[page=1]{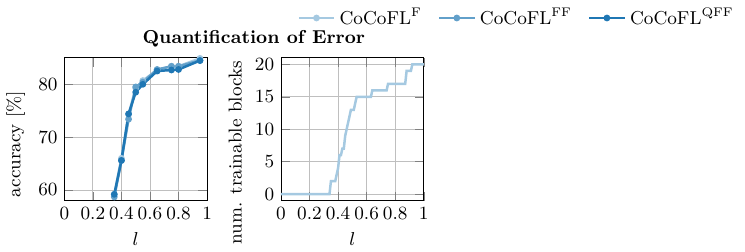}
       }
       \caption{(Left) final accuracy not utilizing gains (error analysis), (right) cumulative number of trainable blocks used for training. When ignoring the gains, the introduced errors from quantization and fusion are negligible.}
       \label{fig:ablation_study2}
  \end{subfigure}
    \begin{subfigure}[t]{0.49\textwidth}
     \trimbox{0.0cm 0.0cm 0.0cm 0.0cm}{
        \includegraphics[page=1]{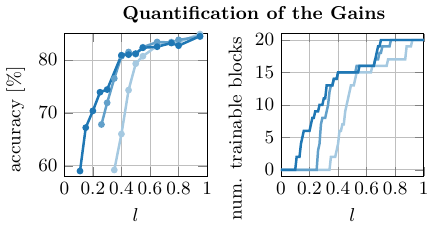}
        }
         \caption{(Left) final accuracy utilizing the gains, (right) cumulative number of trainable blocks used for training. When utilizing the gains, \emph{limited} devices can train more blocks and achieve higher accuracies, outweighing the error.}
         \label{fig:ablation_study}
  \end{subfigure}
  \caption{Quantification of the gains and the error.}
\end{figure}

We conduct an ablation study to quantify the gains and the error of quantization and operator fusion of frozen blocks in CoCoFL. For this purpose, we modify the MobileNet/CIFAR10 \ac{IID} experiment of \cref{subsec:evaluation_setup}.
Instead of three, we have two groups:~10\% \emph{strong} devices, i.e., no constraints (training the full \ac{NN}). We label the remaining~90\% \emph{limited} devices, with a computation and memory limit~$l$, s.t.~$t_\text{limited}(A)=\frac{1}{l}\cdot t_\text{strong}(A)$ and~$M_{\text{limited}} = l \cdot M_{\text{strong}}$.
We apply no communication constraint, therefore, $S_{\text{limited}} = S_{\text{strong}}$. Consequently, a \emph{limited} device has to select a configuration~$A$ that satisfies both the computation and the memory constraint. Several experiments are conducted, where~$l$ is varied between~$l \in [0,1]$. The remaining hyperparameters are kept the same (\cref{tab:hyperparameters}). We introduce three variants of CoCoFL, where we profile each variant's configurations~$A$ to measure the execution time and the memory footprint:
\begin{itemize} [noitemsep, nolistsep]
    \item \cocoflf, where only freezing and no quantization or operator fusion is applied. The set of feasible configurations is denoted as~$\hat{\mathcal A}^{\text{F}}$.
    \item \cocoflff is a variant where freezing and fusion of operators, but no quantization is applied. Configurations are denoted as~$\hat{ \mathcal A}^{\text{FF}}$.
    \item \cocoflqff is the mainline variant (\cref{subsec:evaluation_setup}). The configurations are equivalent to~$\hat{\mathcal{A}}$.
\end{itemize}
\textbf{Quantification of the Error:} To quantify the error introduced through fusion and estimation of the statistics as well as quantization noise, we run all three variants \cocoflf, \cocoflff, and \cocoflqff for different values of~$l$, but each variant uses the configurations in~$\hat{\mathcal{A}}^{\text{F}}$ on the \emph{limited} devices (ignoring all computation/memory gains that come from quantization and fusion). For a given~$l$, all variants can train exactly the same configurations, hence, the same number of blocks.
This allows studying the introduced errors independently of the gains in performance/memory.
On the right part of \cref{fig:ablation_study2}, the cumulative number of trainable blocks for a given~$l$ in $\hat{\mathcal{A}}^{\text{F}}$ is displayed.
Using $\hat{\mathcal{A}}^{\text{F}}$, at least~$l=0.35$ is required to train a single block. The accuracy results are visualized in \cref{fig:ablation_study2} (left), where the  accuracy for different values of~$l$ is reported on the left. Overall, the error is mostly below~$\SI{1}{\pp}$, with a maximum of $\SI{2.3}{\pp}$ for \cocoflqff and $\SI{1.7}{\pp}$ for \cocoflff. Note that this analysis ignores the performance and memory gains, which are studied in the next section.

\textbf{Quantification of the Gains:} To quantify the gains, we run the experiments with all three variants where each variant uses its own profiling results. Hence, \cocoflqff and \cocoflff use~$\hat{\mathcal{A}}$, and~$\hat{\mathcal{A}}^{\text{FF}}$, respectively. We measure that \cocoflqff reaches a maximum reduction of 75\% of computation time and 60\% of memory w.r.t.~to \cocoflf (45\% and 28\% with \cocoflff). Therefore, at a given constraint~$l$, \cocoflqff and \cocoflff can train more configurations and hence, more importantly, configurations with more trained blocks.
This results in an overall higher accuracy in \ac{FL} as can be seen in \cref{fig:ablation_study} (left), where with the same constraint of~$l=0.25$ \cocoflqff achieves an increase of the final accuracy of~\SI{6.5}{\pp} over \cocoflff, while \cocoflf has no configuration on the \emph{limited} devices that satisfies the constraint. At a constraint of~$l=0.4$, \cocoflqff achieves a final accuracy increase of~\SI{14.9}{\pp} over \cocoflf, while \cocoflff achieves a final accuracy increase of~\SI{14.3}{\pp} over \cocoflf. From the results it can be concluded that the more blocks the \emph{limited} devices can train, the higher the final accuracy. This is visualized in \cref{fig:ablation_study} (right), where cumulatively the total number of blocks that can be trained for a given constraint is plotted. The figures also show that in the case of the constraint approaching~$l=1.0$, the advantage of quantization and fusion is vanishing, and can even result in small accuracy losses due to the introduced error.

\emph{In summary, for limited devices, the benefits of fusion and quantization of blocks, i.e., training more blocks with the same available resources, largely outweigh the introduced error. Only when the devices' constraint approaches full resources, the gains do vanish. Overall, quantization and fusion increase the \ac{FL} system's accuracy, as limited devices can make a higher contribution to the model.}

\section{Conclusion}
We proposed CoCoFL that is able to better incorporate knowledge from constrained devices into the \ac{FL} model, especially in non-\ac{IID} cases, preserving fairness among participants. Our comparison with the state of the art, based on real hardware measurements, shows that CoCoFL reaches significantly higher final accuracies. We believe that the gains through quantization can even be higher on devices like smartphones that have on-chip integer \ac{NN} accelerators.

In an \ac{FL} system, devices are acquiring data through sensing or interaction with the environment. As devices are distributed in the system, they may have access to different types of data. Examples include sensors that sample environments that differ from each other, or smartphones that interact with users with different behaviors. This is therefore important to guarantee that we learn from all these devices, regardless of their capabilities, as any piece of the gathered data matters. What is then important is to provide fairness (accuracy parity) among devices, fairness of participation alone, as was the focus of state of the art, is not enough. By approaching accuracy parity among devices, CoCoFL makes \ac{FL} systems applicable to a broader range of use cases, especially use cases when the distribution of classes across devices is skewed.

\subsubsection*{Acknowledgments}
This work is in parts funded by the Deutsches Bundesministerium
für Bildung und Forschung (BMBF, Federal Ministry of Education
and Research in Germany). The authors acknowledge support by
the state of Baden-Württemberg through bwHPC.

\bibliography{bib/references.bib}
\bibliographystyle{tmlr}

\newpage

\appendix

\section{Profiling Setup Details}
\label{sec:appdx_profiling}

\textbf{Profiling}: Profiling of the used \acp{NN} is done for CoCoFL as well as HeteroFL and FjORD to quantify the reduction in training time, maximum memory usage, and upload volume. Two platforms are used: An x86-64 AMD Ryzen~7 with 64\,GB RAM and a Raspberry Pi~4 (Cortex-A72 ARM v8 64-bit) with 8\,GB RAM. The profiling is done for all tested \ac{NN} architectures to acquire all configurations in~$\hat{\mathcal{A}}$. Similarly, we profile HeteroFL and FjORD where we vary the drop ratio from~$0.1 - 1.0$ with~$50$ linearly spaced steps. We allow freezing of the last layer (linear layer)~$b_N$ but do not apply quantization. Profiling takes (MobileNet) $\SI{17}{\minute}$ on x64 and about~$\SI{1}{\hour}$ on ARM. The profiling procedure measures the following quantities in the training:
\begin{itemize} [noitemsep, nolistsep]
    \item Maximum memory: The following training-related parts are included in the memory measurements: First, the model is loaded from disk, second, a training batch is loaded from the disk, and third, an optimizer is initiated. These operations are followed by training of~$16$ mini-batches. The maximum memory is measured by using the Linux syscall \texttt{getrusage()}, with parameter \texttt{RUSAGE\_SELF}. This call returns a struct with a variable \texttt{ru\_maxrss}, that stores the maximum amount of memory, the process required at some point in time. For all measurements, we subtract the maximum memory before the training (e.g., the overhead of the process, PyTorch, and NumPy imports).
    \item Training time: For the same training procedure, we measure the training time. The time measurements include the \ac{NN}'s forward pass execution, setting gradients to zero, the backpropagation step, and optimizer steps. We do not account for mini-batch-level switching between configurations for FjORD and model the switching as a zero-overhead operation.
    \item Upload volume: The upload volume can be directly calculated from the size of the \ac{NN}s' \texttt{state\_dicts} by filtering for parameters that required gradient calculations during training.
\end{itemize}
\textbf{SOTA Comparison:} For HeteroFL and FjORD we enforce the same constraints from \cref{eq:constraint} as for CoCoFL. To ensure this we configure the baselines in the following way:
\begin{itemize} [noitemsep, nolistsep]
    \item FjORD~\citep{horvath2021fjord}: In FjORD all devices switch between different configurations for each mini-batch. We use FjORD's drop levels of~$\mathcal{P}^{\text{FjORD}} =\{20\%,40\%,\ldots,100\%\}$. A device~$c$ with lower resources can only train with levels~$p_j$, that satisfy the constraint~$\mathcal{P}_c^{\text{FjORD}} = \{ p_j\,|\,t_c(p_j) \leq T \land m_c(p_j) \leq M_c \land S(p_j) \leq S_c\}$.
    \item HeteroFL~\citep{diao2020heterofl}: Similarly, in HeteroFL the ratio of dropped filters in an \ac{NN} is set through a shrinkage ratio~$k \in (0,1]$, where~$k=1$ results in training of all available filters. Contrary to FjORD, a device uses the same shrinkage ratio throughout the round. For each device $c$, we set the shrinkage ratio to the maximum value that satisfies the resource constraints~$k_c =
    \max_{0<k\le1} \,\,\text{s.t.}\,\, t_c(k) \leq T \land m_c(k) \leq M_c \land S(k) \leq S_c$.
\end{itemize}

\section{Implementation Details and Hyperparameters}
\label{sec:appdx_details}
\textbf{Miscellaneous Hyperparameters for Vision Models}: We evaluate our technique in an \ac{FL} setup where we train the \ac{NN} models DenseNet40~\citep{densenet}, MobileNetV2~\citep{howard2017mobilenets}, ResNet50~\citep{gao2020residual}, and ResNet18~\citep{gao2020residual}. The image datasets CIFAR10~\citep{krizhevsky2009learning}, CINIC10~\citep{darlow2018cinic}, CIFAR100~\citep{krizhevsky2009learning}, XChest~\citep{Wang_2017_CVPR} and FEMNIST~\citep{cohen2017emnist} are used, where each individual image has a resolution of~$32\times32$ pixels, with~$3$ color channels. In the case of FEMNIST, we scale the~$28\times28$ grayscale image to~$32\times32$ and~$3$ channels to have the same \ac{NN} structure, independent of the dataset type. Additionally, we do not split the written digits and numbers by writers, as proposed by \cite{caldas1812leaf}, but randomly distribute the images to the devices in case of \ac{IID} to have an equal amount of data on each device. A mini-batch size of~$32$ is used for all experiments. For XChest~\citep{Wang_2017_CVPR} we sample $12.7$K samples from the full available dataset and train for finding/no finding. The images are downscaled to $256\times256$ with $3$ color channels. Per round, each active device~$c \in \mathcal{C}^{(r)}$ trains for one local epoch. We apply no data augmentation techniques.

\textbf{Miscellaneous Hyperparameters for NLP Models:} We evaluate our technique in an FL setup using Transformers~\citep{vaswani2017attention} with two datasets. In the case of IMDB~\citep{imdb} a \emph{sentencepiece} tokenizer is used with a vocabulary size of $16,000$ to detect if a movie review is positive or negative. In the case of Shakespeare~\citep{caldas1812leaf} every character of the alphabet represents a possible token. For Transformer models the baselines HeteroFL and FjORD scale down the feature embedding and hidden size instead of training with subsets of CNN filters. To adjust to the resource requirements we use $[12.5\%, 40\%, 62.5\%, 81.25\%, 100\%]$ of the hidden/embedding dim, yet, for IMDB a large memory overhead remains, hence, \emph{weak} devices have to be dropped from the training.

\textbf{Centralized Experiments}: For centralized experiments, we reduce the number of rounds $R$ by a factor of 10, hence, per experiment, we train for $\frac{R}{10}$ epochs over the full dataset. We adjust the learning rate decay steps accordingly.

\section{Configuration Selection Ablation Study}
\label{appdx:block_selection_ablation_study}
To verify the robustness of our configuration selection heuristic~(\cref{sec:technique_block_selection}), specifically, the per-device per-round random selection of a configuration out of $\mathcal{A}_{\text{max}}$, we perform several experiments.  We reuse the setting from the ablation study (\cref{sec:ablation_study}) using MobileNet with CIFAR10, $10$\% \emph{strong} devices and~$90$\% \emph{limited} devices. The \emph{strong} devices train the \ac{NN} end-to-end. We run several experiments with~$l \in [0.2, 0.3, 0.4, 0.5]$ to verify that our heuristic is robust within a large range of constraints (and available configurations). Firstly, we study the effect of our configuration reduction mechanism. Specifically, we compare:
\begin{itemize}[noitemsep, nolistsep]
\item \textbf{max:} Keeping only maximal configurations $\mathcal{A}_{\text{max}}{\subseteq}\mathcal{A}_f$,i.e., configurations that are not a subset of other feasible configurations (as used in CoCoFL).
\item \textbf{all:} Keeping all feasible configurations $\mathcal{A}_f{\subseteq}\hat{\mathcal{A}}$.
\end{itemize}
We further compare our random approach against other mentioned baselines, such as
\begin{itemize}[noitemsep, nolistsep]
\item \textbf{max}: Using the configuration that trains the maximum number of blocks within the device's capabilities. The combination of this selection mechanism and both reduction mechanisms (\textbf{max+max} and \textbf{max+all}) result in the same configurations selected, hence, we only evaluate it once.
\item \textbf{min}: Training the configuration with the minimum number of blocks.
\item \textbf{round-robin}: Switching between feasible configurations in a round-based manner (all limited devices train the same configuration in a round).
\item \textbf{random}: Randomly switching between configurations (as used in CoCoFL).
\end{itemize}

\begin{figure}[t]
    \centering
    \includegraphics[page=1]{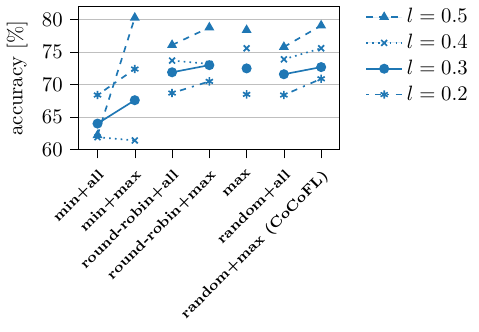}
    \caption{Configuration selection heuristic ablation study evaluating with $l \in [0.2,0.3,0.4,0.5]$.}
    \label{fig:heuristic_ablation}
\end{figure}

We provide the average final accuracy of three independent runs after $1000$ rounds in~\cref{fig:heuristic_ablation}. We observe that in almost all cases, using only maximal configurations (i.e., configurations that are not a subset of another feasible configuration) increases the final accuracy independent of the selection strategy. Further, we observe
that \textbf{random+max} (as done in CoCoFL), independent of~$l$, is within the highest-performing selection strategies. It can be observed from the results that training the same blocks on all devices does not improve upon random, as \textbf{round-robin+max} performs worse than \textbf{random+max}. Depending on~$l$, \textbf{max} and \textbf{min+max} can outperform \textbf{random+max}, but not consistently throughout different values of~$l$.

\end{document}